# Image-Based Model Parameter Optimization Using Model-Assisted Generative Adversarial Networks

Saúl Alonso-Monsalve and Leigh H. Whitehead

*Abstract*— We propose and demonstrate the use of a model-assisted generative adversarial network (GAN) to produce fake images that accurately match true images through the variation of the parameters of the model that describes the features of the images. The generator learns the model parameter values that produce fake images that best match the true images. Two case studies show excellent agreement between the generated best match parameters and the true parameters. The best match model parameter values can be used to retune the default simulation to minimize any bias when applying image recognition techniques to fake and true images. In the case of a real-world experiment, the true images are experimental data with unknown true model parameter values, and the fake images are produced by a simulation that takes the model parameters as input. The model-assisted GAN uses a convolutional neural network to emulate the simulation for all parameter values that, when trained, can be used as a conditional generator for fast fake-image production.

## I. INTRODUCTION

Generative adversarial networks (GANs) [1] have been shown to produce fake images indistinguishable from true images, but these images are manipulated in an arbitrary way to match the true image. However, there are many experiments, for example, in experimental physics, where the features of the true images can be described by a model that contains a number of parameters (hereafter referred to as the model parameters). In these cases, the GAN should only be able to manipulate the images in a way described by the model parameters to produce accurate and physically motivated fake images. Considering a set of true images produced by such an experiment, with fake images produced by an implementation of the model called the simulation, it is clear that an arbitrary approach to manipulating the fake images to match the true images is not well motivated as it disregards any knowledge of the model and the model parameters. Note that in a real experiment, the true images are experimental data images, but, here, we will, henceforth, refer to them as true images.

We propose the model-assisted GAN as a solution to this problem. The approach varies the model parameters $p = (p_0, \ldots, p_N)$ that cause well-defined changes in the images, providing new fake images that better match the true images. A set of true images is produced with true model parameter vectors $p_t$ drawn from a distribution such that $p_t \sim p_{\text{data}}(p_t)$ since a single choice of $p_t$ would produce a set of identical images. The goal is for the model-assisted GAN to generate a set of model parameters $p_{\text{bm}} \sim p_{\text{generator}}(p_{\text{bm}})$ to produce simulated images that best match the true images such that $p_{\text{generator}}(p_{\text{bm}}) = p_{\text{data}}(p_t)$. The default simulation for an experiment will typically have default model parameter values that do not exactly match the true model parameter values. The parameters $p_{\text{bm}} \sim p_{\text{generator}}(p_{\text{bm}})$ can be extracted to update the default simulation model parameters so that the fake images will more accurately reproduce the true images. Furthermore, the difference between $p_{\text{generator}}(p_{\text{bm}})$ and the default parameters distribution gives physical insight into the understanding of the model and the model parameter values that were not correct in the default simulation. The key advantage of the model-assisted GAN is that the simulation only needs to be run once, and then, any number of model parameter optimizations can be performed with different true data sets very quickly. Another advantage is that any additional number of fake images can be produced efficiently using the emulator from the model-assisted GAN.

## II. RELATED WORK

To the best of our knowledge, there is no GAN variant in the literature that aims to generate a vector of parameters that are used to produce fake images through a defined mapping of the parameters to an image as opposed to generating the fake images directly. However, there are some related studies to consider. One example is the conditional GAN [2] that was used to generate MNIST digits conditioned on class labels. More recent studies used conditional GANs for more complex tasks, such as generating aged versions of people's faces that preserve their identities [3]. There are also some novel works that propose a conditional GAN framework that is robust against forgetting in generative models [4].

During the last few years, some studies successfully learned knowledge constraints from image and text generation [5] that were used to improve the results over base generative models or to learn disentangled representations in a completely unsupervised manner [6]. In addition, Creswell and Bharath [7] introduced a new inversion technique to identify attributes of a data set that a trained GAN is able to model and quantitatively compare the performance of different generative networks.

Several domains could benefit from the approach we present in this brief, but its best application is probably in physical experiments [8]. Although GANs have not been broadly used in real-world scientific experiments, some promising work has been done on the production of $jet$ images [9], GAN-based calorimeter simulations [10], [11], and in the production of galaxy images [12]. Contrary to the above-mentioned studies, the model-assisted GAN that we present in this brief could be, for instance, used to learn the optimal parameters needed by a Monte Carlo simulation for mimicking detector images in physics experiments.

## III. NETWORK ARCHITECTURE AND DATA TYPES

A schematic of the model-assisted GAN architecture is shown in Fig. 1. The three data types (true, simulated, and emulated), the four neural networks (generator, emulator, discriminator, and Siamese), and the simulator that forms the architecture are described in the following. The training and implementation of the network are also discussed.

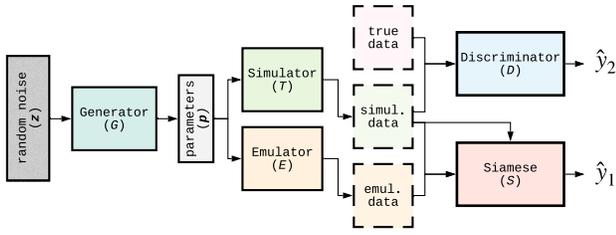

Fig. 1. Overview of the model-assisted GAN.

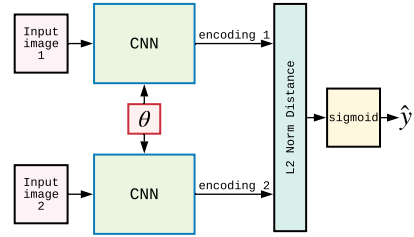

Fig. 2. Architecture of the Siamese network, containing two identical CNNs that share the same parameters $\theta$, used to determine the similarity of two input images.

### A. Data Types

There are three $m \times m$ pixel image data types.

1) *True Data:* A set of simulated data with a chosen set of true parameters $p_t$ used to represent an experiment. In a real experiment, this would be the experimental data with unknown parameter values that we want to measure.
2) *Simulated Data:* The output from the custom simulation used to simulate the experiment.
3) *Emulated Data:* The output from the emulator that learns to mimic the custom simulation.

### B. Generator

The goal of the generator $G$ is to produce parameters such that the simulator can use them to create images that cannot be distinguished from true data images. Contrary to the generator from a traditional GAN, this generator outputs only the set of parameters $p$ rather than the completed image. These parameters form the input to both the simulator and the emulator. In our implementation, $G$ is a multilayer perceptron (MLP), and its last layer uses a tanh activation function.

### C. Simulator

The simulator $T$, which is specific to each use case, can be any set of operations that perform a well-defined transformation from the input model parameters $p$ to an $m \times m$ pixel image. The simulator is not a neural network (or any other type of machine learning algorithm) but, for example, a Monte Carlo simulation of an experiment with some default set of model parameter values that do not necessarily produce fully accurate fake images due to uncertainties associated with the understanding of the experiment. Two simulation choices are used in the case studies described in Sections IV and V.

### D. Emulator

The emulator $E$ is a neural network, similar to a generator from a conditional GAN [2], that learns to mimic the simulator. In other words, its aim is to generate identical images to those of the simulator $T$ when both $E$ and $T$ are fed with the same input parameters $p$. It is a necessary component of the architecture as it provides the correct back-propagation (it is not generally possible to calculate the derivatives of the simulator) needed by the generator to learn the required model parameter variations. Once the components of the model-assisted GAN have been trained to produce the optimal set of parameters, the emulator can be used as a fast simulation technique since it produces an accurate emulation of the full simulation running in the simulator step. The specific architecture of the emulator depends on the complexity of the simulation and the number of parameters. In our implementation, which is identical for both case studies, $E$ is a convolutional neural network (CNN) that maps an input parameter vector $p$ to a 2-D output that represents an emulated image. The last layer of $E$ uses a tanh activation function.

### E. Discriminator

As in regular GANs, the goal of the discriminator $D$ is to distinguish between true data images and images produced by the simulator (or images produced by the emulator to speed up the training process). In our implementation, $D$ is a CNN, and its last layer uses a sigmoid activation function.

### F. Siamese Network

The Siamese network $S$ [13], [14] determines the similarity between the simulated images and the emulated images (both the simulator and the emulator are fed with the parameters $p$ from the generator) and is used to ensure that the images are as identical as possible. It includes two CNNs that share all their parameters so that they give the same output for a given input. For the $i$th training example, each CNN generates an encoding $f(x^{(i)})$ (a 128-length vector) from an input sample $x^{(i)}$; then, the L2-norm of the differences of both encodings is applied (see Fig. 2). Unlike in face recognition deep learning tasks [14], [15], this network outputs the probability of two input images to be identical (same pixel map), so there is no need to use a triplet loss [16], unless the simulator $T$ had the ability to output different images (e.g., by introducing some randomness) from the same parameters $p$. The last layer of the Siamese network uses a sigmoid activation function.

### G. Training Details

The two main stages of training the model-assisted GAN, each consisting of two steps, are shown visually in Fig. 3: pretraining and training.

1) In the pretraining stage, the goal is to learn an emulator distribution $p_E(x)$ that matches the simulator distribution $p_T(x)$. The model learns an emulator network $E$ that generates samples from the emulator distribution $p_E$ by transforming random parameter set variables $p_{\text{random}}(r)$ into samples $E(r)$ such that $E(r) = T(r)$, where $T(r)$ are samples that the custom simulation technique $T$ generates from the simulator distribution $p_T$ by transforming the same random parameter set variables $p_{\text{random}}(r)$ into the samples. To achieve this, $S$ and $E$ play the following two-player minimax game [1] with value function $V_1(E, S)$[1]:

$$\min_E \max_S V_1(S, E) \\ = \mathbb{E}_{r \sim p_{\text{random}}(r)}[\log S(T(r), T(r))] \\ + \mathbb{E}_{r \sim p_{\text{random}}(r)}[\log(1 - S(T(r), E(r)))]. \quad (1)$$

The two pretraining steps using randomized input parameters, normalized between $-1$ and $1$, are as follows.

---
[1]The expectation $\mathbb{E}$, or expected value, of some function $f(x)$ with respect to a probability distribution $p(x)$ is the average, or mean value, that $f$ takes on when $x$ is drawn from $p$ [17].

a) The Siamese network is trained to learn the similarity of the simulated and emulated images.

b) The emulator is trained to learn to create emulated images that mimic simulated images using the Siamese network as trained in step one. The goal is that the emulator and the simulator generate an identical image from all possible parameter sets.

2) In the training stage, the goal is to learn a generator distribution $p_G$ over parameters $\boldsymbol{p}$ such that the parameters $\boldsymbol{p}$ predicted by the generator $G$ can be used by the custom simulation technique $T$ to match the true data distribution $p_{\text{data}}(\boldsymbol{x})$. The generator network $G$ generates parameter samples from the generator distribution $p_G$ by transforming random noise variables $p_{\text{noise}}(z)$ into samples $G(z)$; then, the pretrained network $E$ generates samples from $G(z)$ in the form of $E(G(z))$. To achieve this, $D$ and $G$ play the following two-player minimax game [1] with value function $V_2(G, D)$[1]:

$$\min_G \max_D V_2(D, G)$$
$$= \mathbb{E}_{\boldsymbol{x} \sim p_{data}(\boldsymbol{x})}[\log D(\boldsymbol{x})]$$
$$+ \mathbb{E}_{z \sim p_{noise}(z)}[\log(1 - D(E(G(z))))]. \quad (2)$$

The two training steps are as follows.

a) The discriminator is trained to distinguish between true data images and simulated images. In the standard configuration shown in Fig. 3, $E(G(z))$ in (2) is replaced by $T(G(z))$. In order to speed up the training process and assuming the pretraining ended successfully, the simulated images may be replaced by emulated images without damaging the results.

b) The generator learns to predict model parameters such that the images from the emulator cannot be distinguished from the true data.

The architectures of the networks presented earlier were inspired by DCGANs [18]. The model-assisted GAN has been implemented in Keras [19] on top of Tensorflow [20], and the code is available at https://gitlab.cern.ch/salonsom/model-assisted-gan.

In order to enhance the training phase, we use some suggestions from [21]. We normalize the images between $-1$ and 1 and use a tanh activation function as the last layer of both the generator and the emulator. To stabilize the training and to provide robustness, we use the label smoothing technique described in [22], the label $y$ is set with random values between 0.7 and 1.2 for simulated images and random values from 0.0 to 0.3 for emulated images, and randomly flip a fraction of simulated and emulated image labels when training the different networks [23]. We use the stochastic gradient descent (SGD) optimizer [24] for training the discriminator and the Siamese and the Adam optimizer [25] for training the generator and the emulator as suggested in [18].

## IV. CASE STUDY 1: FIRST-ORDER POLYNOMIAL IMAGES

Consider an experiment that produces $28 \times 28$ pixel images with a signal consisting of a first-order polynomial with gradient $m$, constant $c$, and extent in the $x$-direction $x_{\text{steps}}$ defined by the simple equation: $y = mx + c$.

The model parameters describing the images are $\boldsymbol{p} = (m, x_0, c, x_{\text{steps}})$, where $x_0$ is the initial $x$ value, and the effect of each parameter on the simulated images is easily understood. We choose a set of true parameter values $\boldsymbol{p}_t$ drawn from the Gaussian distributions with means $\mu_i$ and standard deviation $\sigma_i$, where $i$ is the index of the parameter in the range 0 to 3. The ranges of the parameters are used to represent variations from different processes that can lead an experiment to have image-to-image variations in the data sample. The mean and standard deviation of the parameters chosen for the true data images are listed in Table I.

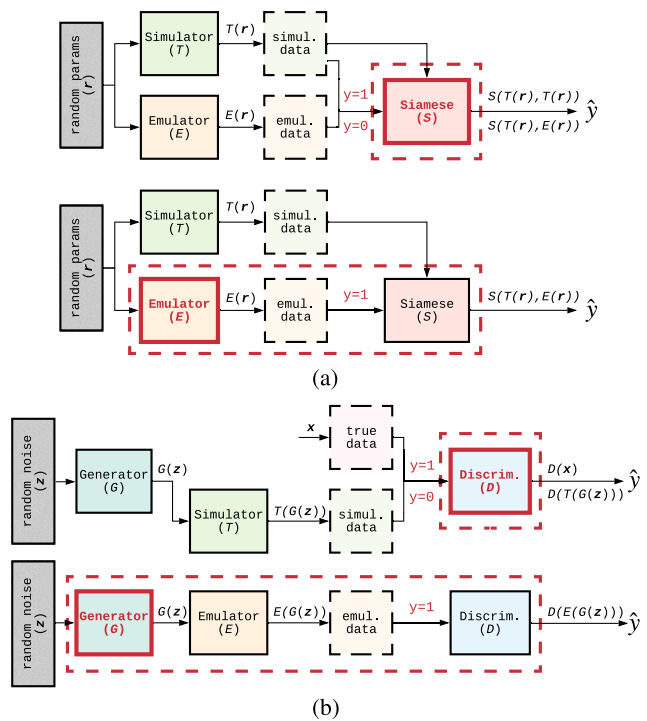

Fig. 3. Two training phases of the model-assisted GAN. The red dashed lines enclose the neural network that is trained in each step, and the inner red boxes correspond to the trainable modules within the neural network (thus, the weights of the other modules, if any, are frozen). (a) Adversarial pretraining: The Siamese network learns the similarity between the simulator and the emulator images; the emulator learns to make emulated data to mimic simulated data. (b) Adversarial training: The discriminator learns to distinguish true (or experimental) data from simulated data (or emulated data to speed up the training); the generator learns to predict parameters $\boldsymbol{p}$ to produce emulated data that best match the true data.

TABLE I
MEAN AND STANDARD DEVIATION OF THE MODEL PARAMETERS USED TO MAKE THE TRUE DATA IMAGES AND THOSE FROM THE BEST MATCH PARAMETERS LEARNED BY THE GENERATOR IN THE FIRST CASE STUDY

| Parameter | $m$ | $x_0$ | $c$ | $x_{steps}$ |
|---|---|---|---|---|
| True $\mu$ | 1.5 | 10.0 | 0.5 | 9.0 |
| True $\sigma$ | 0.3 | 0.5 | 0.1 | 0.5 |
| Best match $\mu$ | 1.501 | 9.988 | 0.512 | 9.003 |
| Best match $\sigma$ | 0.266 | 0.418 | 0.129 | 0.432 |

The model-assisted GAN was trained with a minibatch size of 256 on a single 16-GB NVIDIA Tesla V100 GPU. The pretraining stage was trained for 500k iterations, and the training stage for 30k iterations. The best match parameters $\bar{\boldsymbol{p}}_{\text{bm}}$, defined as the mean of the best match parameter values from the generator, and the corresponding standard deviations are shown in the bottom two rows of Table I to be in very good agreement with the true data parameters.

Example images from the simulator and emulator are shown in Fig. 4 for five random parameter sets after 1k, 5k, 10k, 100k, and 500k pretraining iterations and demonstrate the ability of the emulator to mimic the simulator. Fig. 5 shows the simulator images at five stages throughout the training process. In each case, three simulated images from randomly chosen sets of generated parameters are shown and compared with three randomly selected true images. These simulated images should not be identical to the true images as they do not have identical parameters, but they are representative of the sample.

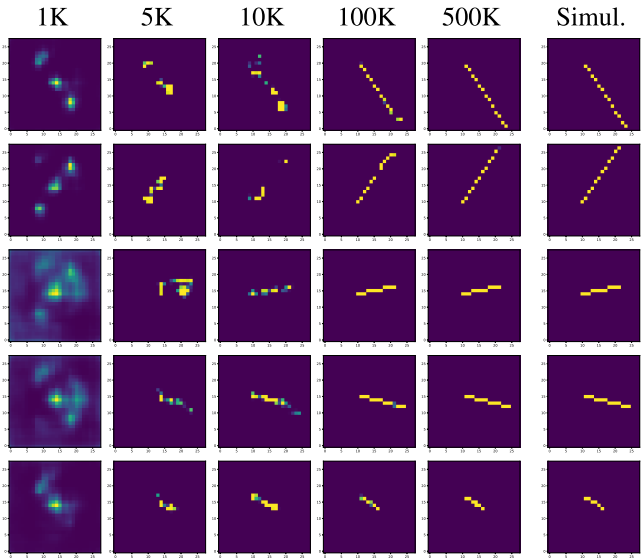

Fig. 4. Emulator output (the columns 1–5 show images after 1k, 5k, 10k, 100k, and 500k pretraining iterations, respectively) versus simulator output (column 6). The rows show the images generated from five arbitrary parameter sets.

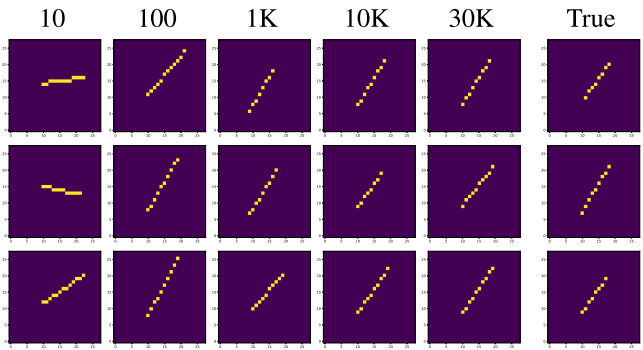

Fig. 5. Generator–simulator output from five random noise vectors (the columns 1–5 show images after 10, 100, 1k, 10k, and 30k training iterations, respectively) versus some random true images (column 6).

## V. CASE STUDY 2: CIRCULAR SIGNAL WITH NOISE AND AMPLITUDE VARIATION

We now consider a more complex example containing model parameters that include brightness and noise manipulations as well as topological changes to $28 \times 28$ pixel images. The topological description of the signal in the images is

$$x^2 + y^2 = r^2 \qquad (3)$$

where $r$ is the radius of the circle. The five model parameters are $\boldsymbol{p} = (x_0, y_0, r, n, b)$, where $(x_0, y_0)$ is the center of the circle, $n$ is the white noise scale that varies the brightness of the white noise, and $b$ is the signal brightness. Both $n$ and $b$ are defined as a fraction of the maximum image brightness. The set of true data model parameter values $\boldsymbol{p}_t$ is produced by drawing from the Gaussian distributions for each model parameter, as described in Section IV. The mean and standard deviation of the parameter values chosen for the true data images are given in Table II.

We used the same training and testing infrastructure described in Section IV. Since the simulator is more complex than in case study one, 1M iterations were needed in the pretraining step. The training step required only 30k iterations, as before.

TABLE II
MEAN AND STANDARD DEVIATION OF THE MODEL PARAMETERS USED TO PRODUCE THE TRUE DATA IMAGES AND THE CORRESPONDING VALUES FROM THE BEST MATCH PARAMETERS FROM THE GENERATOR IN THE SECOND CASE STUDY

| Parameter | $x_0$ | $y_0$ | $r$ | $n$ | $b$ |
|---|---|---|---|---|---|
| True $\mu$ | 18 | 18 | 8 | 0.15 | 0.9 |
| True $\sigma$ | 0.8 | 0.8 | 0.5 | 0.05 | 0.05 |
| Best match $\mu$ | 18.082 | 18.035 | 7.950 | 0.152 | 0.892 |
| Best match $\sigma$ | 0.833 | 0.795 | 0.411 | 0.072 | 0.083 |

TABLE III
IMAGE PRODUCTION TIME COMPARISON OF THE SIMULATOR AND THE EMULATOR FOR EACH CASE STUDY. EACH VALUE SHOWS THE AVERAGE TIME OF TEN EXECUTIONS

| Image source | | | Number of generated images | | |
|---|---|---|---|---|---|
| | | | $10K$ | $100K$ | $1M$ |
| Case Study 1 | Simulator[1] | | 2.143 s | 21.368 s | 213.798 s |
| | Emulator | CPU[1] | 6.539 s | 66.536 s | 672.633 s |
| | | GPU[2] | 1.625 s | 4.901 s | 36.656 s |
| Case Study 2 | Simulator[1] | | 47.367 s | 471.842 s | 4,773.626 s |
| | Emulator | CPU[1] | 6.540 s | 66.541 s | 671.628 s |
| | | GPU[2] | 1.623 s | 4.907 s | 36.671 s |

The mean and standard deviation of the best match parameters from the trained generator are shown in Table II in comparison to the true data parameters. Excellent agreement is seen for all five parameters showing that the model-assisted GAN performs equally well on this more complex example.

Fig. 6 shows a comparison of the emulator images for five sets of arbitrarily chosen parameters compared at 12 points in the pretraining process to the simulated image with the same parameters. The emulated images are shown to accurately reproduce the simulated images in this more complex scenario. Fig. 7 shows three randomly chosen simulated images from different points in the training stage compared with three randomly chosen true images.

## VI. FAST SIMULATION WITH THE EMULATOR

Once trained, the emulator produces images very similar to the simulation in considerably less time for the same set of input model parameters $\boldsymbol{p}$. Table III shows that in both the case studies, the emulator running on the GPU is much faster than the simulation and that the emulator execution time is independent of the complexity of the simulation for a given emulator architecture. The emulator is also considerably quicker using the CPU for case study two, and it is only slower for the first case study since the simulation is very simple, which will not be the case for a real-world experiment. The emulator could, hence, be used in place of the simulation to allow for the rapid development of analyses that can be performed without needing the exact simulated images.

## VII. CONCLUSION

We have proposed and demonstrated the use of model-assisted GANs to produce physically motivated manipulations of the simulated images through variation of underlying model parameters. In two case studies presented here, the model-assisted GAN produces best match parameters $\boldsymbol{p}_{\text{bm}} \sim p_{\text{generator}}(\boldsymbol{p}_{\text{bm}})$ in excellent agreement with the true data parameters $\boldsymbol{p}_t \sim p_{\text{data}}(\boldsymbol{p}_t)$ and, hence, generates simulated and emulated images that accurately match the true images. In a situation with experimental data instead of true data, the best match parameters $p_{bm}$ would be used to update the default simulation to produce more accurate images that reproduce the experimental data images. This is critical to minimize biases and ensure the similar performance of image-recognition techniques applied to simulated images and data images in experimental situations.

[2]The experiments were run on an Intel Xeon CPU E5-2698 v4 @ 2.20 GHz.
[3]The experiments were run on a 16GB NVIDIA Tesla V100 GPU.

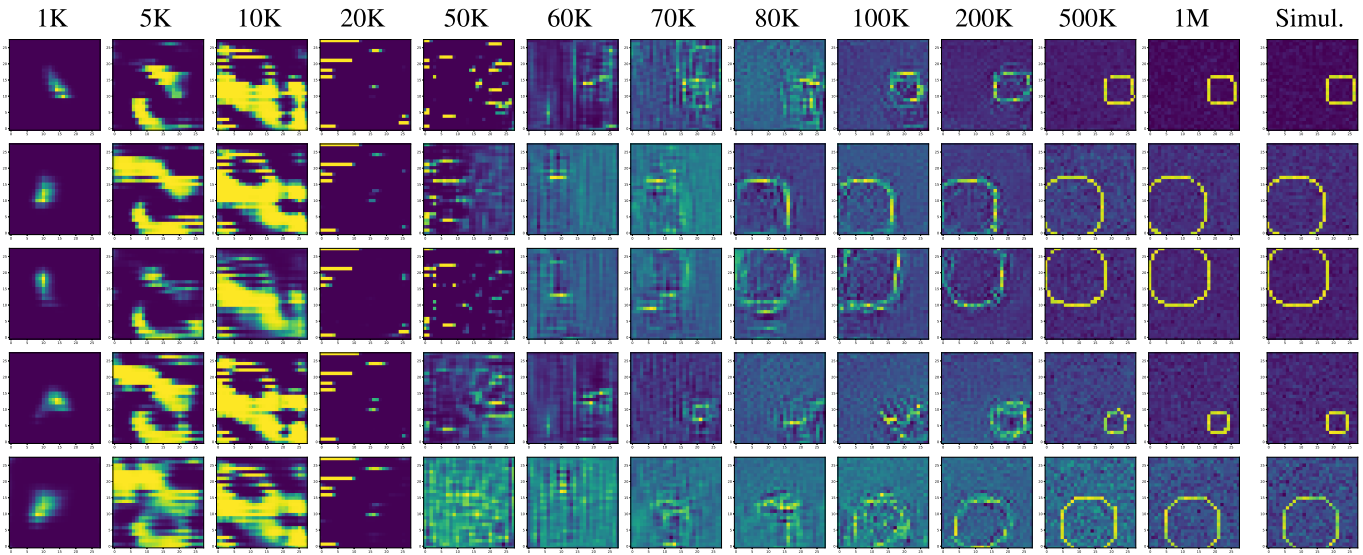

Fig. 6. Emulator output (the columns 1–12 show images after 1k, 5k, 10k, 20k, 50k, 60k, 70k, 80k, 100k, 200k, 500k, and 1M pretraining iterations, respectively) versus simulator output (column 13). The rows show the images generated from five arbitrary parameter sets.

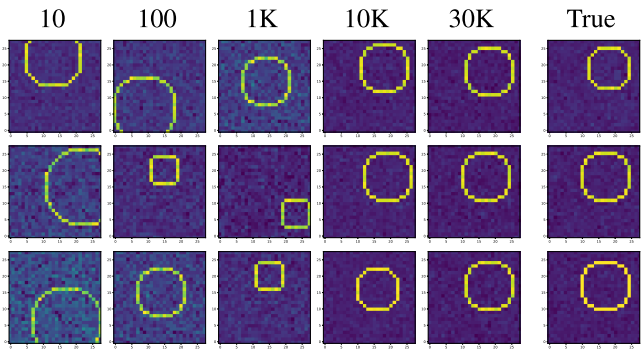

Fig. 7. Generator–simulator output from some random noise vectors (the columns 1–5 show images after 10, 100, 1k, 10k, and 30k training iterations, respectively) versus some random true images (column 6).

The emulator that is trained as a part of the model-assisted GAN can be used as a conditional generator to very quickly produce images very similar to the simulation for a given set of model parameters. The advantages of this method of image production become increasingly clear for complex simulations, and the first step of the training shown in Fig. 3(b) could use the emulator and emulated data in order to reduce the training time accumulated from using a very complex simulation.

In the future, we will explore fully realistic applications of the model-assisted GAN to real experiments in a number of scientific disciplines, including high-energy physics, for both simulation parameter optimization and fast simulations.

ACKNOWLEDGMENT

The authors would like to thank Prof. A. Aurisano for the provision of computing resources at the University of Cincinnati.

REFERENCES

[1] I. Goodfellow *et al.*, "Generative adversarial nets," in *Proc. Adv. Neural Inf. Process. Syst.*, 2014, pp. 2672–2680.

[2] M. Mirza and S. Osindero, "Conditional generative adversarial nets," 2014, *arXiv:1411.1784*. [Online]. Available: https://arxiv.org/abs/1411.1784

[3] G. Antipov, M. Baccouche, and J.-L. Dugelay, "Face aging with conditional generative adversarial networks," 2017, *arXiv:1702.01983*. [Online]. Available: https://arxiv.org/abs/1702.01983

[4] C. Wu *et al.*, "Memory replay GANs: Learning to generate new categories without forgetting," in *Proc. Adv. Neural Inf. Process. Syst.*, 2018, pp. 5962–5972.

[5] Z. Hu *et al.*, "Deep generative models with learnable knowledge constraints," 2018, *arXiv:1806.09764*. [Online]. Available: http://arxiv.org/abs/1806.09764

[6] X. Chen, Y. Duan, R. Houthooft, J. Schulman, I. Sutskever, and P. Abbeel, "InfoGAN: Interpretable representation learning by information maximizing generative adversarial nets," in *Proc. Adv. Neural Inf. Process. Syst.*, 2016, pp. 2172–2180.

[7] A. Creswell and A. A. Bharath, "Inverting the generator of a generative adversarial network," *IEEE Trans. Neural Netw. Learn. Syst.*, vol. 30, no. 7, pp. 1967–1974, Jul. 2019.

[8] A. Radovic *et al.*, "Machine learning at the energy and intensity frontiers of particle physics," *Nature*, vol. 560, no. 7716, pp. 41–48, Aug. 2018.

[9] L. de Oliveira, M. Paganini, and B. Nachman, "Learning particle physics by example: Location-aware generative adversarial networks for physics synthesis," *Comput. Softw. Big Sci.*, vol. 1, no. 1, p. 4, Sep. 2017.

[10] M. Paganini, L. de Oliveira, and B. Nachman, "CaloGAN: Simulating 3D high energy particle showers in multilayer electromagnetic calorimeters with generative adversarial networks," *Phys. Rev. D, Part. Fields*, vol. 97, no. 1, Jan. 2018, Art. no. 014021.

[11] M. Paganini, L. de Oliveira, and B. Nachman, "Accelerating science with generative adversarial networks: An application to 3D particle showers in multilayer calorimeters," *Phys. Rev. Lett.*, vol. 120, no. 4, Jan. 2018, Art. no. 042003.

[12] K. Schawinski, C. Zhang, H. Zhang, L. Fowler, and G. K. Santhanam, "Generative adversarial networks recover features in astrophysical images of galaxies beyond the deconvolution limit," *Monthly Notices Roy. Astronomical Soc., Lett.*, vol. 467, no. 1, pp. L110–L114, 2017.

[13] J. Bromley *et al.*, "Signature verification using a 'siamese' time delay neural network," *Int. J. Pattern Recog. Artif. Intell.*, vol. 7, no. 4, pp. 669–688, 1993.

[14] S. Chopra, R. Hadsell, and Y. LeCun, "Learning a similarity metric discriminatively, with application to face verification," in *Proc. IEEE Comput. Soc. Conf. Comput. Vis. Pattern Recognit. (CVPR)*, vol. 1, Jun. 2005, pp. 539–546.

[15] Y. Taigman, M. Yang, M. Ranzato, and L. Wolf, "DeepFace: Closing the gap to human-level performance in face verification," in *Proc. IEEE Conf. Comput. Vis. Pattern Recognit.*, Jun. 2014, pp. 1701–1708.

[16] F. Schroff, D. Kalenichenko, and J. Philbin, "FaceNet: A unified embedding for face recognition and clustering," 2015, *arXiv:1503.03832*. [Online]. Available: http://arxiv.org/abs/1503.03832

[17] I. Goodfellow, Y. Bengio, and A. Courville, *Deep Learning*. Cambridge, MA, USA: MIT Press, 2016. [Online]. Available: http://www.deeplearningbook.org

[18] A. Radford, L. Metz, and S. Chintala, "Unsupervised representation learning with deep convolutional generative adversarial networks," 2015, *arXiv:1511.06434*. [Online]. Available: https://arxiv.org/abs/1511.06434

[19] F. Chollet *et al.* (2015). *Keras*. [Online]. Available: https://github.com/keras-team/keras

[20] M. Abadi *et al.*, "Tensorflow: A system for large-scale machine learning," in *Proc. OSDI*, vol. 16, 2016, pp. 265–283.

[21] S. Chintala, E. Denton, M. Arjovsky, and M. Mathieu. (2016). *How to Train a GAN? Tips and Tricks to Make GANs Work*. [Online]. Available: https://github.com/soumith/ganhacks

[22] T. Salimans, I. Goodfellow, W. Zaremba, V. Cheung, A. Radford, and X. Chen, "Improved techniques for training GANs," 2016, *arXiv:1606.03498*. [Online]. Available: https://arxiv.org/abs/1606.03498

[23] M. Arjovsky and L. Bottou, "Towards principled methods for training generative adversarial networks," 2017, *arXiv:1701.04862*. [Online]. Available: https://arxiv.org/abs/1701.04862

[24] L. Bottou, F. E. Curtis, and J. Nocedal, "Optimization methods for large-scale machine learning," *SIAM Rev.*, vol. 60, no. 2, pp. 223–311, Jan. 2018.

[25] D. P. Kingma and J. Ba, "Adam: A method for stochastic optimization," 2014, *arXiv:1412.6980*. [Online]. Available: https://arxiv.org/abs/1412.6980